\def\year{2021}\relax
\documentclass[letterpaper]{article} %
\usepackage{aaai21}  %
\usepackage{times}  %
\usepackage{helvet} %
\usepackage{courier}  %
\usepackage[hyphens]{url}  %
\usepackage{graphicx} %
\usepackage{natbib}  %
\usepackage{caption} %
\usepackage{makecell}
\usepackage{microtype}
\usepackage{multirow}
\usepackage{engord}
\usepackage[switch]{lineno}

\newcommand*\samethanks[1][\value{footnote}]{\footnotemark[#1]}

\title{Benchmarking Knowledge-Enhanced Commonsense Question Answering \\via Knowledge-to-Text Transformation}
\author {
    Ning Bian,\textsuperscript{\rm 1,3}
    Xianpei Han,\textsuperscript{\rm 1,2,\thanks{Corresponding authors.}}
	Bo Chen,\textsuperscript{\rm 1,2}
    Le Sun\textsuperscript{\rm 1,2,\samethanks[1]} \\
}
\affiliations {
    \textsuperscript{\rm 1}Chinese Information Processing Laboratory,\ 
    \textsuperscript{\rm 2}State Key Laboratory of Computer Science \\
	Institute of Software, Chinese Academy of Sciences, Beijing, China \\
    \textsuperscript{\rm 3}University of Chinese Academy of Sciences, Beijing, China\\
  \{bianning2019, xianpei, chenbo, sunle\}@iscas.ac.cn
}

\begin{document}
\maketitle

\begin{abstract}
A fundamental ability of humans is to utilize commonsense knowledge in language understanding and question answering. In recent years, many knowledge-enhanced Commonsense Question Answering (CQA) approaches have been proposed. However, it remains unclear: (1) How far can we get by exploiting external knowledge for CQA? (2) How much potential of knowledge has been exploited in current CQA models? (3) Which are the most promising directions for future CQA? To answer these questions, we benchmark knowledge-enhanced CQA by conducting extensive experiments on multiple standard CQA datasets using a simple and effective knowledge-to-text transformation framework. Experiments show that: (1) Our knowledge-to-text framework is effective and achieves state-of-the-art performance on CommonsenseQA dataset, providing a simple and strong knowledge-enhanced baseline for CQA; (2) The potential of knowledge is still far from being fully exploited in CQA — there is a significant performance gap from current models to our models with golden knowledge; and (3) Context-sensitive knowledge selection, heterogeneous knowledge exploitation, and commonsense-rich language models are promising CQA directions.
\end{abstract}

\section{Introduction}
\begin{figure}
   \centering
   \includegraphics[width=\columnwidth]{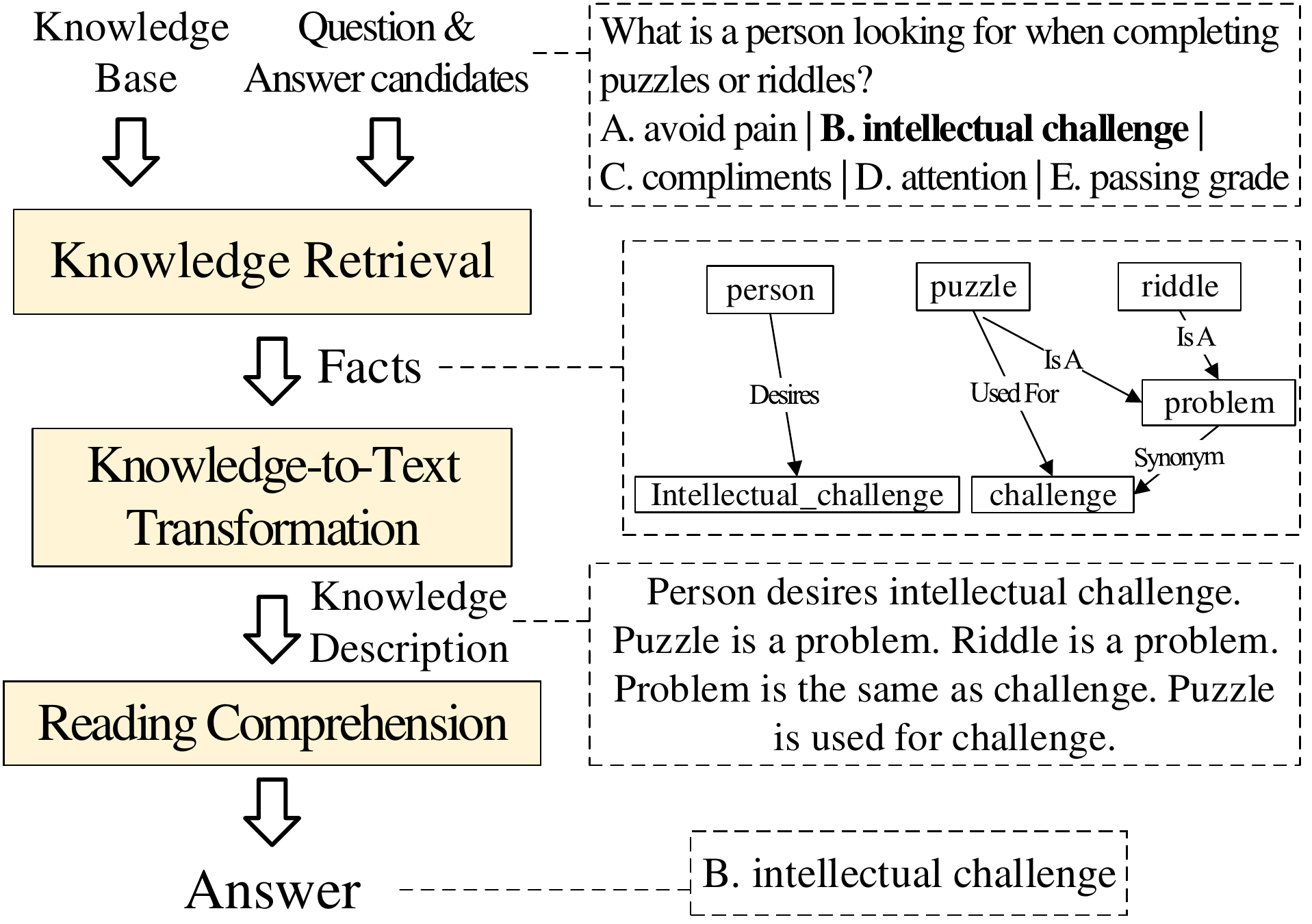}
   \caption{Our knowledge-to-text framework for benchmarking knowledge-enhanced CQA with an example from CommonsenseQA \cite{talmor2019commonsenseqa}.}
   \label{figure-f1}
\end{figure}

\noindent Using a variety of knowledge to help in understanding the meaning of language is one of the key abilities of humans \cite{minsky2000commonsense}. Commonsense question answering (CQA) evaluates whether machines can understand language like humans do by asking questions whose answers rely on commonsense knowledge. For example, Figure \ref{figure-f1} shows a question, and the answer to this question needs commonsense knowledge “\textit{puzzle is used for intellectual challenge}”. 

Witnessed the importance of commonsense knowledge for CQA, many studies have been conducted to incorporate external knowledge bases (KBs) in CQA models. These approaches usually leverage knowledge to enhance a specific CQA component: 1) enhancing representations \cite{weissenborn2017dynamic,bauer-etal-2018-commonsense,mihaylov-frank-2018-knowledgeable,ma2019towards}; 2) enhancing attention mechanism \cite{chen-etal-2018-neural-natural,wang-jiang-2019-explicit}; and 3) enhancing reasoning mechanism \cite{lin-etal-2019-kagnet,lv2020graph}.

Although many knowledge-enhanced CQA approaches have been proposed, we found it is still unclear: (1) How far can we get by exploiting external knowledge for CQA? (2) How much potential of knowledge has been exploited in current models? For example, can GNN-based models \cite{lin-etal-2019-kagnet,lv2020graph} encode and exploit all useful evidence provided by external knowledge? (3) Which are the most promising directions for knowledge-enhanced CQA? We believe answering these questions can provide valuable insights for future CQA studies and shed light on other knowledge-dependent tasks like reading comprehension \cite{rajpurkar-etal-2016-squad} and conversation generation \cite{zhou2018commonsense}.

To answer the above questions, we benchmark knowledge-enhanced CQA by conducting extensive experiments on multiple standard datasets via a simple and effective knowledge-to-text transformation framework. Intuitively, to benchmark knowledge-enhanced CQA, external knowledge should be incorporated in a simple way that is not specialized to specific models/components. This is challenging, due to 1) the heterogeneity between structured knowledge and unstructured textual questions/answers, i.e., knowledge facts are usually triples such as $<$\textit{person, \textbf{Desires}, Intellectual\_challenge}$>$, but questions and answers are text; and 2) the context-sensitivity of knowledge, i.e., a KB may contain thousands of facts about a concept, but only several of them are relevant to the given question. For example, among the thousands of facts about “\textit{person}”, only $<$\textit{person, \textbf{Desires}, Intellectual\_challenge}$>$ is useful for answering the question in Figure \ref{figure-f1}.

Specifically, our knowledge-to-text framework consists of three stages, which are shown in Figure \ref{figure-f1}. Firstly, we retrieve facts from a commonsense knowledge graph (CKG). Then we transform the knowledge facts to textual descriptions via three transformation algorithms (template-based, paraphrasing-based, and retrieval-based). Finally, we utilize machine reading comprehension (MRC) models to predict answers by exploiting both the original questions and the textural knowledge descriptions. This framework is simple and general for benchmarking knowledge-enhanced CQA: 1) By transforming structured knowledge into textual descriptions, our method resolves the heterogeneity problem between knowledge and text. 2) By adopting MRC models, our method can learn to select question-relevant knowledge automatically. 3) Our simple knowledge-enhancing strategy allows us to easily compare the effects of different commonsense knowledge. 

We conduct thorough experiments on multiple standard CQA datasets \cite{talmor2019commonsenseqa,levesque2012winograd,zellers-etal-2019-hellaswag,sap-etal-2019-social}. 

The contributions of our paper are:

1. Through benchmarking experiments we found that the potential of external knowledge is still far from exploited in knowledge-enhanced CQA, i.e., current methods can only exploit knowledge to a limited extent. In our experiments, there is a big performance gap from current models to our models using golden knowledge.

2. We propose a simple and effective knowledge-to-text framework for knowledge-enhanced CQA which achieves state-of-the-art performance on the CommonsenseQA dataset, providing a simple and strong knowledge-enhanced baseline for CQA.

3. Our experimental results shed light on three important future directions for knowledge-enhanced CQA: context-sensitive knowledge selection, heterogeneous knowledge exploitation, and commonsense-rich language models.

\section{Knowledge-enhanced CQA via Knowledge-to-Text Transformation}
Following CommonsenseQA \cite{talmor2019commonsenseqa}, the CQA task in this paper is a multiple-choice problem with five answer candidates. Given question $Q=[q_1,q_2,…,q_n ]$ and answer candidates $A=\{A_1,A_2,…,A_m \}$ with each answer candidate $A_k=[a_1^k,a_2^k,…,a_l^k ]$,  $a_j^k$ and $q_i$ are words, $i$ and $j$ are indexes of words and $k$ is the index of answer candidate, a CQA model needs to choose the correct answer from $A$. 

We propose a simple and effective knowledge-to-text framework for benchmarking knowledge-enhanced CQA. Our framework includes three steps: 1) retrieving facts from CKG; 2) transforming knowledge to text; and 3) adopting an MRC model to select the answer. 

Notice that the purpose of our paper is to benchmark knowledge-enhanced CQA rather than to propose new techniques. So, it is critical to select classical, robust, and well-known models, rather than new models which may lead to biased conclusions. Our framework is not specialized to a specific CQA setting, therefore it can also be used in other MRC or QA tasks. 

In the following, we describe the three stages of our framework. 

\subsection{Knowledge Retrieval}
To answer a question $Q$, our method first retrieves relevant knowledge from a given CKG. For example, to answer the question in Figure \ref{figure-f1}, we want to retrieve facts like $<$\textit{person, \textbf{Desires}, Intellectual\_challenge}$>$ and $<$\textit{puzzle, \textbf{UsedFor}, challenge}$>$. Following a previous study \cite{lin-etal-2019-kagnet}, we retrieve paths on CKG connecting question concepts and answer concepts as relevant facts, which provides a good precision/recall trade-off for question-relevant facts.

Concretely, given a question $Q$ and an answer candidate $A_k$, we first identify concepts in them by exactly matching n-grams with the concepts in CKG. (we use ConceptNet \cite{speer2017conceptnet} in this paper). Then, for each pair of $<$\textit{question concept, answer candidate concept}$>$, we find all paths between them on CKG (within $K$ hops) as facts for $A_k$ ($K$ is a hyper-parameter here). For the example in Figure \ref{figure-f1}, “\textit{puzzle$\to$\textbf{IsA}$\to$problem$\to$\textbf{Synonym}$\to$challenge}” is a 2-hop knowledge path for answer candidate “\textit{intellectual challenge}”.

\subsection{Knowledge-to-Text Transformation}
This section describes how to resolve the heterogeneity problem between knowledge and text via knowledge-to-text transformation. Specifically, we propose three transformation algorithms: template-based, paraphrasing-based, and retrieval-based, which are described as follows.

\renewcommand{\arraystretch}{1.3}
\begin{table*}[!]
	\small
\begin{center}
    \begin{tabular}{p{5.2cm}m{3.39cm}m{3.6cm}m{3.9cm}}
    \hline \makecell[c]{\textbf{Knowledge path}} & \makecell[c]{\textbf{Template-based}} & \makecell[c]{\textbf{Paraphrasing-based}} & \makecell[c]{\textbf{Retrieval-based}} \\
    
	\hline \makecell[c]{Silk$\to$\textbf{AtLocation}$\to$China} & Silk is located in China. & Silk is in China. & China is the world's largest silk producer. \\

    \hline \makecell[c]{Puzzle$\to$\textbf{IsA}$\to$Problem$\to$\textbf{Synonym}\\$\to$Challenge} & \textls[-20]{Puzzle is a problem. Problem is the same as challenge.} & \textls[-50]{Puzzles are problems. The problem is the same as the challenge.} & Puzzle problem is a challenge game for children. \\
   
    \hline \makecell[c]{Walk$\gets$\textbf{MotivatedByGoal}$\gets$Hike$\to$\textbf{Have-} \\\textbf{Subevent}$\to$See beautiful views} & \textls[0]{Hike in order to walk. Hike have subevent see beautiful views.} & \textls[-20]{You go hiking in order to go for a walk. You can see the beautiful scenery on hiking.} & \textls[-30]{Burghclere has some beautiful rural scenery, so you can walk along the railway or go for a hike.} \\
    \hline
    \end{tabular}
	\end{center}
\caption{\label{table1} Examples of knowledge descriptions generated by different knowledge-to-text algorithms.}
\end{table*}%

\textbf{Template-based transformation.} This algorithm transforms knowledge to text using a description template for each relation in a CKG. For example, we can use a template “X is a Y” to generate the description of $<$\textit{puzzle, \textbf{IsA}, problem}$>$ as “\textit{puzzle is a problem}”. Because the number of relations in a CKG is limited, we manually design a template for each relation type. For a knowledge path $\{k_1,k_2,…,k_p,… \}$ where $k_p$  is a knowledge triple and $p$ is its index, we sequentially generate a sentence for each tuple, i.e., $\{s_1,s_2,…,s_p  ,… \}$ where sentence $s_p$ describes triple $k_p$.

\textbf{Paraphrasing-based transformation.} The main drawback of the template-based algorithm is the diversity issue, i.e., it always generates the same description for one relation. To address this issue, we employ a paraphrasing model to generate more diverse and fluent knowledge descriptions. Specifically, given the template-based description of a knowledge path, we generate its top-$M$ paraphrases using beam-search decoding and concatenate them as the knowledge description. We adopt the encoder-decoder paraphrasing model trained on PPDB \cite{pavlick-etal-2015-ppdb} and WikiAnswers \cite{fader-etal-2013-paraphrase}.

\textbf{Retrieval-based transformation.} The above two algorithms can only generate pseudo textual descriptions, which are different from real-world knowledge descriptions. Therefore, we propose a retrieval-based knowledge-to-text algorithm, which retrieves texts from a real-world corpus (we use Wikipedia in this paper) as knowledge descriptions. Specifically, we adopt the distant supervision assumption \cite{mintz-etal-2009-distant} that “\textit{if a sentence contains the entities on a knowledge path, it will express the meaning of the knowledge path}”. We split all Wikipedia documents into separate sentences and build a Wikipedia sentence retrieval system using Elastic Search. We use the knowledge descriptions from template-based transformation as queries to retrieve corresponding Wikipedia sentences containing the concepts on knowledge paths via the BM25 algorithm \cite{robertson1994some}. Finally, the rank 1 sentence is used as the description.

To compare different knowledge-to-text transformation algorithms, Table \ref{table1} shows some examples of generated knowledge descriptions. We can see that: (1) The template-based algorithm can produce reasonable textual descriptions, although they may contain grammar errors (like “\textit{Hike in order to walk}” in the \engordnumber{3} example). (2) The paraphrasing-based algorithm can produce diverse and more fluent sentences (“\textit{You go hiking in order to go for a walk}”), but may change some important words (e.g., “\textit{beautiful view}” is changed to “\textit{beautiful scenery}” in the \engordnumber{3} example). (3) The retrieval-based algorithm can produce real-world sentences (“\textit{China is the world's largest silk producer}”) but may contain extra irrelevant content (like “\textit{Burghclere}” in the \engordnumber{3} example). 

\subsection{MRC-based Answer Prediction}
Given a question and the generated knowledge descriptions, we predict its answer using MRC models. We adopt MRC models because: 1) MRC models can automatically learn to identify relevant information in a document \cite{seo2016bidirectional}. In our settings this ability can be used to automatically select question-relevant knowledge, as all knowledge facts have been transformed into a textual document; 2) MRC is a well-studied technique. Therefore, our method can directly leverage the strong ability of existing state-of-the-art MRC models, so that our benchmarking is effective, robust, and easy-to-implement.

\begin{figure}
   \centering
   \includegraphics[width=\columnwidth]{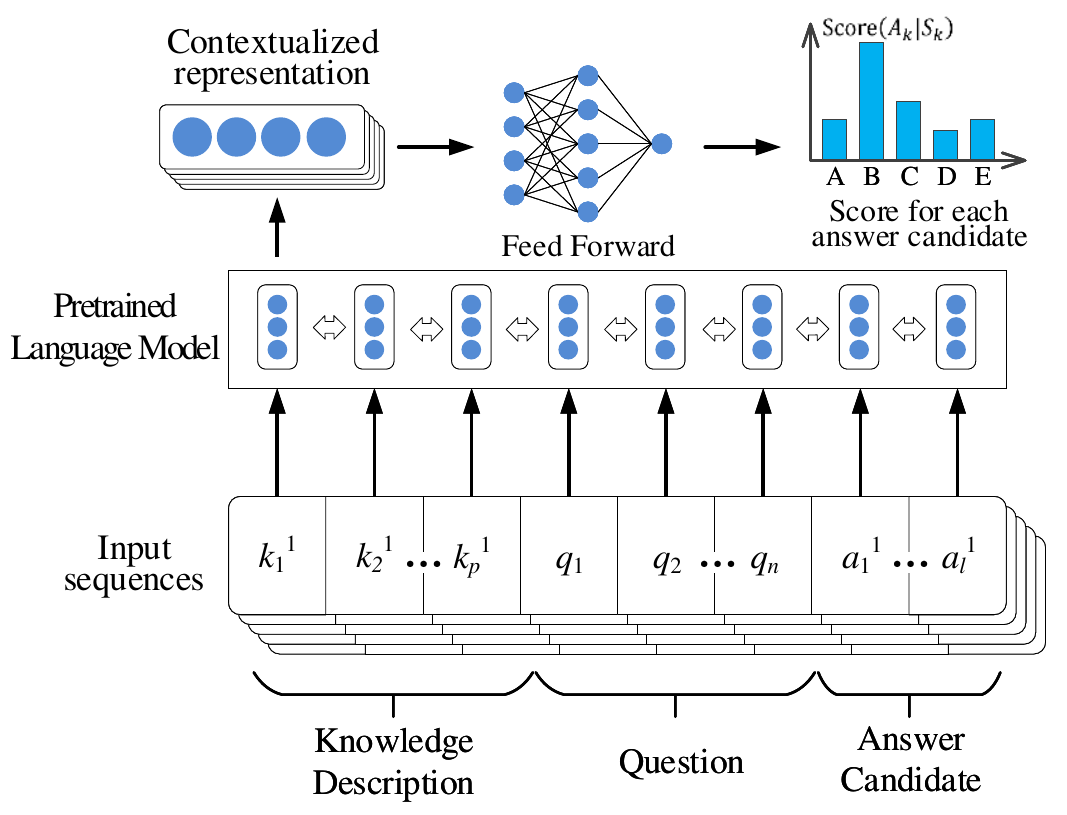}
   \caption{The MRC model for predicting answers in our knowledge-enhanced CQA method.}
   \label{figure-f2}
\end{figure}

Specifically, we model CQA as an MRC problem by treating knowledge descriptions as a document. In this way, current MRC models can be directly used, including BERT \cite{devlin-etal-2019-bert}, RoBERTa \cite{liu2019roberta}, XLNet \cite{yang2019xlnet}, and ALBERT \cite{lan2019albert} based MRC models. Figure \ref{figure-f2} shows our MRC framework. For each question, we construct a sequence $S_k=\{K_k,[SEP],Q,[SEP]  ,A_k \}$ for each  answer candidate $A_k$, where $K_k$ is the generated knowledge descriptions, $Q$ is the question, and $[SEP]$ is the separation token in pretrained language models (PLMs). Following \citet{devlin-etal-2019-bert}, we use a feed-forward classifier as the output layer which predicts the answer score $\rm{Score}$$(A_k |S_k )$. Finally, the highest-scored answer candidate is chosen as the answer.

\section{Benchmarking Knowledge-Enhanced Commonsense Question Answering}
This section benchmarks knowledge-enhanced CQA by conducting thorough experiments. We first verify the effectiveness and robustness of our knowledge-to-text-based CQA method, then we answer the three important questions: (1) \textit{How far can we get by exploiting external knowledge for CQA?} (2) \textit{How much potential of knowledge has been exploited in current models?} (3) \textit{Which are the most promising directions for future knowledge-enhanced CQA?}

\renewcommand{\arraystretch}{1.1}
\begin{table*}[t!]
\begin{center}

    \begin{tabular}{lccccc}
	\hline
    \textbf{Model} & \textbf{Knowledge Source} & \textbf{BERT} & \textbf{XLNet} & \textbf{RoBERTa} & \textbf{ALBERT} \\
    \hline
    \textbf{Human} & --    & 88.9  & 88.9  & 88.9  & 88.9 \\
    \hline
    \textbf{Golden Knowledge} & Human Explanations & 81.1  & 85.1  & 84.7  & 83.7 \\
    \hline
    \textbf{Knowledge-to-Text} &  &       &       &       &  \\
    \quad Template-based & ConceptNet & 67.9  & 77.5  & 78.1  & 81.1 \\
    \quad Paraphrasing-based & ConceptNet & 67.2  & 74.9  & 77.8  & 79.3 \\
    \quad Retrieval-based & ConceptNet & 65.0    & 75.0    & 77.1  & 79.4 \\
    \quad Full  & ConceptNet & \textbf{70.4} & \textbf{80.3} & \textbf{80.8} & \textbf{83.3} \\
    \hline

    \makecell[l]{\textbf{\textls[-20]{Best Knowledge-enhanced}} \\\textbf{\textls[-20]{System with Different PLMs}}} & ConceptNet & \makecell[c]{69.0\\\textls[-30]{\cite{ma2019towards}}}    & \makecell[c]{79.3\\\textls[-30]{\cite{lv2020graph}}}  & \makecell[c]{80.8\\(KEDGN)}  & \makecell[c]{\textls[-20]{(No available} \\\textls[-20]{model so far)}} \\

    \hline
    \textbf{Base Model} & No knowledge & 63.6  & 68.9  & 76.2  & 78.6 \\
    \hline
    \end{tabular}%
\end{center}
\caption{\label{table2} Accuracies on CommonsenseQA. }
\end{table*}%

\subsection{Experimental Settings}
\textbf{Datasets.} We use CommonsenseQA dataset v1.11 \cite{talmor2019commonsenseqa} as the primary dataset, and adopt the Winograd Schema Challenge (WSC, \citeauthor{levesque2012winograd} \citeyear{levesque2012winograd}), HellaSWAG \cite{zellers-etal-2019-hellaswag}, and SOCIAL IQa \cite{sap-etal-2019-social} as secondary datasets. 

(1) \textit{CommonsenseQA} \cite{talmor2019commonsenseqa} contains 12,102 human-generated questions with 5 answer candidates for each question. All questions are elaborately designed to make sure commonsense knowledge is needed for correctly answering them. Furthermore, CoS-E \cite{rajani-etal-2019-explain} provides each question with a human-annotated golden knowledge explanation. Due to the above advantages, We use CommonsenseQA as the primary benchmarking dataset.

(2) \textit{WSC} \cite{levesque2012winograd} is a pronoun resolution dataset that requires commonsense knowledge, which is recognized as one of the most difficult CQA datasets \cite{zhou2020evaluating}. Because WSC does not contain training data, we use WSCR \cite{rahman-ng-2012-resolving} for training.

(3) \textit{HellaSWAG} \cite{zellers-etal-2019-hellaswag}  is an update of the commonsense reasoning dataset SWAG: given an event description like “\textit{A woman sits at a piano}”, a machine needs to select the most likely follow-up: “\textit{She sets her fingers on the keys}”. The “Overall accuracy” on the dev set is used in our evaluation.

(4) \textit{SOCIAL IQa} \cite{sap-etal-2019-social} is a QA dataset for commonsense reasoning about social situations, which requires emotional and social commonsense in a variety of every-day situations.

\textbf{Knowledge base.} We use ConceptNet 5 \cite{speer2017conceptnet} as the KB for benchmarking, because: (i) ConceptNet is general and can provide a large commonsense coverage for our CQA experiments. Other CKGs like ATOMIC (\citeauthor{sap2019atomic} \citeyear{sap2019atomic}, if-then relations of events) and ASER (\citeauthor{zhang2020aser} \citeyear{zhang2020aser}, relations of events, states, and actions) only contain partial knowledge for our experiments. (ii) The primary CommonsenseQA dataset is constructed upon ConceptNet and other datasets don't accompany a given KB. ConceptNet concepts can be easily and directly identified in questions and answers for CommonsenseQA, so that we can better benchmark knowledge-enhanced CQA by focusing on the ability of knowledge exploitation. We use the same 22 relations in ConceptNet as  \citet{talmor2019commonsenseqa}.

\textbf{Baselines.} We benchmark knowledge-enhanced CQA by assessing the performances of different MRC models with/without external knowledge, including BERT-based \cite{devlin-etal-2019-bert}, RoBERTa-based \cite{liu2019roberta}, XLNet-based \cite{yang2019xlnet}, and ALBERT-based \cite{lan2019albert} MRC models.

To verify the effectiveness of knowledge-to-text transformation, we also report the performances of current knowledge-enhanced systems with corresponding pretrained language models as base encoders: 

(1) \citet{ma2019towards} (BERT + OCN + ConceptNet) is the best BERT-based knowledge-enhanced CQA system on CommonsenseQA, which uses an attention mechanism for knowledge incorporation and an Option Comparison Network (OCN) model for answer prediction. 

(2) \citet{lv2020graph} (XLNet + Graph Reasoning) is the best XLNet-based system on CommonsenseQA, which uses GNN to exploit knowledge from both ConceptNet and Wikipedia.

(3) KEDGN (RoBERTa + Knowledge) is the unpublished best RoBERTa-based knowledge-enhanced system on the leaderboard of CommonsenseQA, which exploits knowledge via a dual graph network. For a fair comparison, in Table \ref{table2} we report the accuracy of the best single model as described in its report.

\textbf{Hyperparameters.} For knowledge retrieval, we use knowledge paths within 2 hops ($K$ = 2). In paraphrasing-based transformation, we use the top 1 paraphrasing result ($M$ = 1). For MRC models, we initialize them with the official pretrained language models (BERT-Large, RoBERTa-Large, XLNet-Large, and ALBERT-XXLarge) and fine-tune them using CQA training data. The output layers have a 1024-dimensional hidden layer with a $tanh$ activation function. All models are trained using Adam with a learning rate of 5e-6.

\subsection{Effect of Knowledge-to-Text Transformation}
Table \ref{table2} and Table \ref{table3} show the experimental results on CommonsenseQA and other datasets. For our method, we use four settings: template-based, paraphrasing-based, retrieval-based, and a full model that uses a concatenation of all the three generated descriptions as a document. We found that:

1) \textit{Knowledge-to-text transformation is effective for knowledge-enhanced CQA.} Our full model achieves state-of-the-art performance on CommonsenseQA. And all template-based, paraphrasing-based, and retrieval-based models achieve improvements over non-knowledge base models.

2) \textit{Knowledge-to-text transformation can robustly exploit knowledge for CQA.} Table \ref{table3} shows that our method can consistently improve the performances on three extra CQA datasets by exploiting external commonsense knowledge. Notice that although ConceptNet is not specially designed for WSC, HellaSWAG, and SOCIAL IQA datasets, our method can still achieve improvements, which further verifies the robustness of our method, and we believe the results on these datasets can be further improved if more relevant commonsense knowledge sources are available. In Table \ref{table2} our method achieves accuracy improvements on all base models (BERT, RoBERTa, XLNet, and ALBERT) and all settings (template-based, paraphrasing-based, and retrieval-based). Table \ref{table4} shows that our method is also robust on different lengths of knowledge paths, and the 2-hop knowledge path setting achieves the best performance.

3) \textit{The three knowledge-to-text transformation algorithms are complements of each other.} In Table \ref{table2}, the full model can achieve the best performance by combining all three knowledge-to-text algorithms, which verifies that these algorithms can complement each other. Among the three single algorithms, the template-based algorithm obtains the best performance. This may be because it is easier for MRC models to capture regularities in simple and formal sentences.

Overall, the above results verify that our simple knowledge-to-text transformation is a good strategy for benchmarking the effectiveness and robustness of knowledge-enhanced CQA. 

In the following, we conduct benchmarking experiments on the primary CommonsenseQA dataset using the full model and 2-hop knowledge path setting.

\begin{table}[tp!]
\begin{center}
  \small
    \begin{tabular}{l|ccc}
    \hline
    \textbf{Models} & \multicolumn{1}{m{1.3cm}}{\makecell[c]{\textbf{WSC}}} & \textbf{HellaSWAG} & \textbf{SOCIAL IQa} \\
    \hline
    BERT  & 66.0    & 42.3  & 66.2 \\
    \ + Knowledge & \textbf{68.1} & \textbf{44.2} & \textbf{68.8} \\
    \hline
    RoBERTa & 81.4  & 82.5  & 74.3 \\
    \ + Knowledge & \textbf{82.5} & \textbf{83.0} & \textbf{75.0} \\
    \hline
    ALBERT & 84.9  & 86.1  & 77.2 \\
    \ + Knowledge & \textbf{87.0} & \textbf{86.9} & \textbf{77.8} \\
    \hline
    Human & 92.1  & 94.5  & 86.9 \\
    \hline
    \end{tabular}%
\end{center}
\caption{\label{table3} Accuracies on other CQA datasets. “+ Knowledge” means using our knowledge-to-text transformation method (template-based) with 2-hop knowledge paths on ConceptNet. Human accuracy of WSC is reported by  \citet{bender2015establishing}.}
\end{table}%

\begin{table}[t!]
  \centering

    \begin{tabular}{ccccc}
    \hline
    \textls[-50]{\textbf{Path Length}} & \textbf{BERT} & \textls[-50]{\textbf{XLNet}} & \textls[-50]{\textbf{RoBERTa}} & \textls[-50]{\textbf{ALBERT}} \\
    \hline
    1-hop & 67.1  & 74.7  & 77.9  & 80.0 \\
    2-hop & \textbf{67.9} & \textbf{77.5} & \textbf{78.1} & \textbf{81.1} \\
    3-hop & 65.0    & 68.6  & 77.2  & 79.2 \\
    \hline
    \end{tabular}%
\caption{\label{table4} Accuracies on different lengths of knowledge paths (template-based method).}
\end{table}%

\begin{table*}[t!]
  \centering
	\small
    \begin{tabular}{m{1.8cm}|p{3.2cm}p{11.0cm}}
    \hline
    \multirow{4}{*}{\makecell[l]{\textbf{Missing} \\\textbf{Important} \\\textbf{Evidence}}} & Question & \textit{What could people do that involves \textbf{talking?}} \\
          & Answer candidates & \textit{\textbf{confession} $|$ state park $|$ sing $|$ opera $|$ carnival} \\
          & Golden knowledge & \textit{\textbf{confession} involves \textbf{talking}.} \\
          & Knowledge description & \textit{people is located in confession. people is used for talk.} \\
    \hline
    \hline
    \multirow{5}{*}{\makecell[l]{\textbf{Complicated} \\\textbf{Descriptions}}} & Question & \textit{They were getting ready for a really long \textbf{hike}, he put the food can in his what?} \\
          & Answer candidates & \textit{\textbf{backpack} $|$ make person sick $|$ cabinet $|$ house $|$ recycling center} \\
          & Golden knowledge & \textit{\textbf{backpacks} are used on \textbf{hicks}.} \\
          & Knowledge description & \textit{food can is located in backpack. \textbf{backpack is in the context of sport. hike is in the context of sport}……} \\
    \hline
    \hline
    \multirow{6}{*}{\makecell[l]{\textbf{Noisy} \\\textbf{Knowledge}}} & Question & \textit{Most \underline{people} who are \textbf{family} \underline{like} to greet each other with a what?} \\
          & Answer candidates & \textit{listen to music $|$ have friends $|$ know what ophiolites $|$ \textbf{hug} $|$ apartments} \\
          & Golden knowledge & \textit{people who are family like to hug.} \\
          & Knowledge description & \textit{\underline{person} desire hug. \underline{person} is located in \textbf{family}. kissing have subevent \textbf{hug}. kissing cause \underline{like}. meeting friend have subevent hug. hug in order to love. love is located in family. most \underline{people} desire hug.}\\
    \hline
    \end{tabular}%
\caption{\label{table5} Bad examples of generated knowledge descriptions (template-based) and golden knowledge, where: 1) In the \engordnumber{1} example, the relational knowledge between “\textit{talking}” and “\textit{confession}” is missing in the generated knowledge description because it is not covered by ConceptNet. 2) In the \engordnumber{2} example, knowledge description provides the knowledge about “\textit{backpack}” and “\textit{hike}” using two separate sentences, which is more complicated than golden knowledge and thus puts an extra burden on MRC models. 3) In the \engordnumber{3} example, there are many irrelevant/noisy sentences in knowledge description about unimportant question words (like “\textit{people}” and “\textit{like}”).}
\end{table*}%

\subsection{Effect of Knowledge for CQA}
This section studies “\textit{how far can we get by exploiting external knowledge for CQA?}”. To answer this question, Table \ref{table2} further shows the performances of MRC models using manually-annotated golden knowledge for each question \cite{rajani-etal-2019-explain} as the knowledge description. We can see that:

\textit{By incorporating golden external knowledge, CQA can be significantly improved and can achieve close-to-human performance.} On all BERT, XLNet, RoBERTa, and ALBERT-based MRC models, incorporating golden knowledge can significantly achieve 27\%, 14\%, 11\%, and 7\% accuracy improvements, correspondingly. The best golden-knowledge enhanced system (XLNet + Golden) can achieve 85.1\% accuracy, which is not far from the human accuracy of 88.9\%.

These results show that knowledge can get us quite far, and it is promising to study more effective knowledge-enhanced CQA models.

\subsection{Effect of Knowledge in Current Models}

This section investigates “\textit{how much potential of knowledge has been exploited in current models?}”. From Table \ref{table2}, we can see that:

1) \textit{Current knowledge-enhanced CQA methods only exploit knowledge to a limited extent.} In Table \ref{table2}, we can see that: (i) compared with models using golden knowledge, all knowledge-enhanced CQA models have a big performance gap; and (ii) our simple knowledge-to-text strategy can achieve competitive performance with the complicated GNN-based strategies (KEDGN and XLNet + Graph Reasoning) and Option Comparison Network.

2) \textit{Despite the effectiveness of our method, there is still great potential in generating accurate question-relevant knowledge descriptions.} To show this, Table \ref{table5} shows several bad cases of knowledge descriptions. We can see that, the golden knowledge descriptions are typically simple, relevant, and accurate, while the automatically generated descriptions may miss important evidence (\engordnumber{1} example), be too complicated (\engordnumber{2} example), or contain noisy knowledge (\engordnumber{3} example). Based on these observations, we believe seeking and identifying more accurate question-relevant knowledge can further improve the knowledge exploitation ability of CQA methods.

3) \textit{The commonsense knowledge embedded in current pretrained language models is still not enough for CQA.} In Table \ref{table2}, we can see that there is a significant performance gap between base models without using knowledge and knowledge-enhanced models, although they have been trained using very large text corpus. To further study this, we also experiment using ERNIE \cite{zhang-etal-2019-ernie}, a knowledge-enhanced pretrained language model based on BERT, but the performance is lower than BERT-based models (60.0\% accuracy on CommonsenseQA). We believe this is because ERNIE focuses on entity-centric facts, instead of commonsense. This shows that, although trained on very large text corpus, state-of-the-art pretrained language models still can not encode enough commonsense knowledge.

The above results show that the potential of knowledge is still far from being fully exploited by current knowledge-enhanced CQA methods. This is because of 1) the limited ability of current CQA models to exploit knowledge; 2) the lack of ability to identify accurate question-relevant knowledge; 3) the limited commonsense captured in pretrained language models.

\subsection{Detailed Analysis}

This section analyzes our method in detail.

\begin{figure}[t!]
   \centering
   \includegraphics[width=\columnwidth]{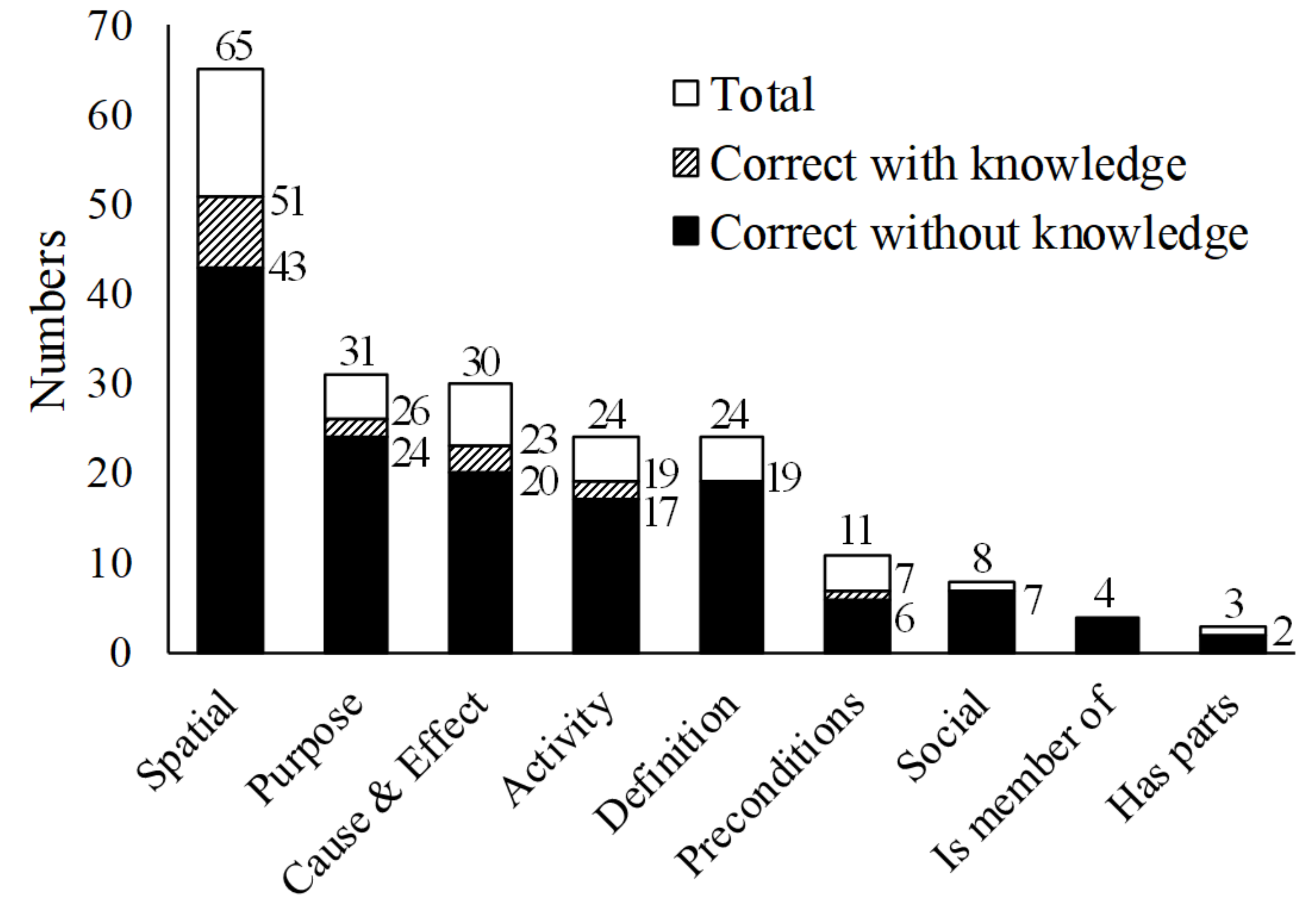}
   \caption{Performances of different commonsense skills using XLNet-based model, with/without knowledge descriptions (template-based).}
   \label{figure-f3}
\end{figure}

\textbf{Performances on Different Commonsense Skills.} CQA questions require different types of commonsense skills \cite{lobue-yates-2011-types}. To analyze the effects of knowledge on different commonsense skills, we randomly sample 200 questions from CommonsenseQA and annotate their required skills using the commonsense skill categories from \citet{talmor2019commonsenseqa}. 

Figure \ref{figure-f3} shows the performances of our CQA method with/without knowledge on different skills. From Figure \ref{figure-f3}, we can see that: (1) Knowledge can significantly improve skills including “\textit{Spatial}” (+12.3\%), “\textit{Cause \& Effect}” (+10.0\%), “\textit{Activity}” (+8.3\%) and “\textit{Purpose}” (+6.5\%). (2) For “\textit{Definition}”, “\textit{Social}”, and “\textit{Has parts}” skills, the knowledge-enhanced model achieves similar performances with the base model. We believe this may be because ConceptNet has a low coverage for these types of knowledge. 

\begin{table*}[t!]
  \centering
    \begin{tabular}{m{2.5cm}|p{4.6cm}p{9.2cm}}
	\hline
    \multicolumn{1}{p{2.5cm}|}{\textbf{Error type}} & \multicolumn{2}{p{13.8cm}}{\textbf{Example}} \\
	\hline
    \multicolumn{1}{p{2.5cm}|}{\multirow{4}[2]{*}{\makecell[l]{\textbf{Indistinguishable} \\\textbf{Knowledge} \\\textbf{(21/50)}}}} & Question & \textit{What do \textbf{airplanes} do as they are arriving at the gate?} \\
          & Answer candidates & \textit{$\surd$ \textbf{slow down} $|$ $\times$ land $|$ $\times$ crash $|$ $\times$ \textbf{speed up} $|$ $\times$ carry people} \\
          & Knowledge for correct answer & \textit{airplanes can slow down.} \\
          & Knowledge for predicted answer & \textit{airplanes can speed up. } \\
    \hline
    \hline
    \multicolumn{1}{p{2.5cm}|}{\multirow{6}[2]{*}{\makecell[l]{\textbf{Noisy Knowledge} \\\textbf{(15/50)}}}} & Question & \textit{I took my \underline{seat}, the \textbf{curtains} drew \underline{back} and I \underline{enjoyed} the what?} \\
          & Answer candidates & \textit{$\times$ auditorium $|$ $\times$ theatre $|$ $\times$ \textbf{movie} $|$ $\surd$ \textbf{show} $|$ $\times$ airplane} \\
          & Knowledge for correct answer & \textit{\textbf{curtain is located in show}. cover is opposite to \underline{back}. \underline{seat} is located in show. show is located in opera. curtain is located in opera. person desire \underline{enjoy}. curtain is located in theater……   } \\
          & Knowledge for predicted answer & \textit{movie is located in theater. curtain is located in theater.                                                                                    } \\
    \hline
    \hline
    \multicolumn{1}{p{2.5cm}|}{\multirow{4}[2]{*}{\makecell[l]{\textbf{No Knowledge}  \\\textbf{(13/50)}}}} & Question & \textit{Some animals can fly thanks to their lightweight hollow what?} \\
          & Answer candidates & \textit{$\times$ heads $|$ $\times$ \textbf{tails} $|$ $\surd$ \textbf{bodies} $|$ $\times$ bones $|$ $\times$ eyes} \\
          & Knowledge for correct answer & \textit{bones is located in person. person desire fly.} \\
          & Knowledge for predicted answer & \textbf{[\textit{NO KNOWLEDGE FACT IS RETRIEVED}]}  \\
    \hline
    \end{tabular}%
\caption{\label{table6} Several error cases of XLNet-based model with template-based knowledge descriptions.}
\end{table*}%

\textbf{Error Analysis.} To understand why our model fails in some cases, we randomly select 50 error cases and group them into several categories. Table \ref{table6} shows the main error types with their examples:

1) \textit{Indistinguishable knowledge}, i.e., retrieved knowledge cannot provide enough information for distinguishing answer candidates. For example, the \engordnumber{1} error case provides strong support for both correct and incorrect answers (“\textit{airplanes can slow down/speed up}”). This is the main error type of our method (21 out of 50).

2) \textit{Noisy knowledge}. Noisy knowledge misleads MRC models to give wrong answers, which often appears when knowledge descriptions are too long. In the \engordnumber{2} error case, we can see that the important fact “\textit{curtain is located in show}” is obscured by noisy facts about irrelevant concepts like “\textit{seat}”.

3) \textit{No Knowledge}. Knowledge retrieval may not be able to retrieve question-relevant facts and thus provides no useful information for MRC models. From the \engordnumber{3} case, we can see that the knowledge facts are all irrelevant to the answers.

The above three types of errors show that it is important to select accurate, complete, and context-sensitive knowledge for more effective knowledge-enhanced models.

\section{Related Work}
\textbf{Knowledge-enhanced CQA.} Many studies have been proposed to exploit commonsense knowledge for CQA.  \citet{rajani-etal-2019-explain} propose to train a GPT-based explanation generation model using manually labeled corpus, but it relies on extra human effort. KagNet \cite{lin-etal-2019-kagnet} represents external knowledge as a graph and reasons via graph convolution and LSTM. \citet{ma2019towards} incorporate knowledge with text-to-knowledge attention and adopt a BERT-based Option Comparison Network for answer prediction. \citet{lv2020graph} propose a GNN-based reasoning model over A heterogeneous knowledge graph of both ConceptNet and Wikipedia sentences. Compared with these methods, our knowledge-to-text method exploits knowledge in a simple way and knowledge can be effectively used by the whole model.

\textbf{Knowledge Exploitation in Neural Models.} There are many studies which leverage external knowledge to enhance models on a variety of NLP tasks \cite{lin-etal-2017-reasoning,yang-mitchell-2017-leveraging,an-etal-2018-accurate,yang-etal-2019-enhancing-pre,logan-etal-2019-baracks,chen-etal-2018-sequence}. \citet{chen-etal-2018-neural-natural} leverage semantic relations in WordNet to enhance attention and inference abilities in the NLI task. \citet{mihaylov-frank-2018-knowledgeable} apply key-value memory to represent commonsense facts and use word-to-knowledge attention for cloze-style MRC. \citet{bauer-etal-2018-commonsense} propose a mutual information-based knowledge selection method and fuse knowledge using gated attention for multi-hop reasoning. \citet{zhang-etal-2019-knowledge} propose an attention-based knowledge selection method for coreference resolution. ERNIE \cite{zhang-etal-2019-ernie} and K-BERT \cite{liu2020k} incorporate knowledge in pretrained language models, but mainly focus on entity-centric facts in KBs instead of commonsense.

\textbf{Machine Reading Comprehension.} In recent years, many effective end-to-end MRC models have been proposed, including BERT \cite{devlin-etal-2019-bert}, RoBERTa \cite{liu2019roberta}, XLNet \cite{yang2019xlnet} and ALBERT \cite{lan2019albert} based models. It has been proven that MRC models can effectively encode information in a document and find the most relevant information for answer prediction. In this paper, these abilities are utilized to select and exploit relevant knowledge for knowledge-enhanced CQA.

\section{Conclusions and Future Work}
We benchmark knowledge-enhanced CQA using a simple and effective knowledge-to-text transformation framework and provides a strong knowledge-enhanced baseline for CQA. By conducting thorough experiments, we found that: (1) Our knowledge-to-text framework is effective and robust for knowledge-enhanced CQA; (2) It is promising to incorporate knowledge in neural models for CQA; (3) The potential of knowledge is still far from being fully exploited — there is a large performance gap from current models to our models using golden knowledge.

The above results also shed light on the promising directions for knowledge-enhanced CQA:

1) \textbf{Context-sensitive knowledge selection is critical for knowledge-enhanced CQA.} According to the error analysis, more than 70\% of errors are caused by noisy knowledge and indistinguishable knowledge.

2) \textbf{The knowledge-text heterogeneity is a critical bottleneck for exploiting the information from both knowledge and text.} We address this heterogeneity problem via simple knowledge-to-text transformation, and even such a simple strategy can outperform many knowledge-enhanced models like GNN-based and attention-based models. Therefore, we believe more advanced solutions for the heterogeneity problem will further improve CQA, e.g., uniform representation learning and joint graph representations.

3) \textbf{It is valuable to incorporate more commonsense in pretrained language models.} From our experiments, we can see that current state-of-the-art pretrained language models like BERT and XLNet still only encode limited commonsense knowledge. So, we believe commonsense-rich language models will provide valuable techniques and resources for CQA.

\section{ Acknowledgments}
This research work is supported by National Key R\&D Program of China under Grant 2018YFB1005100, the National Natural Science Foundation of China under Grants no. U1936207 and 61772505, Beijing Academy of Artificial Intelligence (BAAI2019QN0502), and in part by the Youth Innovation Promotion Association CAS (2018141).

\bibliography{aaai-2021}

\begin{thebibliography}{41}
\providecommand{\natexlab}[1]{#1}
\providecommand{\url}[1]{\texttt{#1}}
\providecommand{\urlprefix}{URL }
\expandafter\ifx\csname urlstyle\endcsname\relax
  \providecommand{\doi}[1]{doi:\discretionary{}{}{}#1}\else
  \providecommand{\doi}{doi:\discretionary{}{}{}\begingroup
  \urlstyle{rm}\Url}\fi

\bibitem[{An et~al.(2018)An, Chen, Han, and Sun}]{an-etal-2018-accurate}
An, B.; Chen, B.; Han, X.; and Sun, L. 2018.
\newblock Accurate Text-Enhanced Knowledge Graph Representation Learning.
\newblock In \emph{Proceedings of the 2018 Conference of the North {A}merican
  Chapter of the Association for Computational Linguistics: Human Language
  Technologies, Volume 1 (Long Papers)}, 745--755. New Orleans, Louisiana:
  Association for Computational Linguistics.

\bibitem[{Bauer, Wang, and Bansal(2018)}]{bauer-etal-2018-commonsense}
Bauer, L.; Wang, Y.; and Bansal, M. 2018.
\newblock Commonsense for Generative Multi-Hop Question Answering Tasks.
\newblock In \emph{Proceedings of the 2018 Conference on Empirical Methods in
  Natural Language Processing}, 4220--4230. Brussels, Belgium: Association for
  Computational Linguistics.

\bibitem[{Bender(2015)}]{bender2015establishing}
Bender, D. 2015.
\newblock Establishing a Human Baseline for the Winograd Schema Challenge.
\newblock In \emph{MAICS}, 39--45.

\bibitem[{Chen, Sun, and Han(2018)}]{chen-etal-2018-sequence}
Chen, B.; Sun, L.; and Han, X. 2018.
\newblock Sequence-to-Action: End-to-End Semantic Graph Generation for Semantic
  Parsing.
\newblock In \emph{Proceedings of the 56th Annual Meeting of the Association
  for Computational Linguistics (Volume 1: Long Papers)}, 766--777. Melbourne,
  Australia: Association for Computational Linguistics.

\bibitem[{Chen et~al.(2018)Chen, Zhu, Ling, Inkpen, and
  Wei}]{chen-etal-2018-neural-natural}
Chen, Q.; Zhu, X.; Ling, Z.-H.; Inkpen, D.; and Wei, S. 2018.
\newblock Neural Natural Language Inference Models Enhanced with External
  Knowledge.
\newblock In \emph{Proceedings of the 56th Annual Meeting of the Association
  for Computational Linguistics}, 2406--2417. Melbourne, Australia: Association
  for Computational Linguistics.

\bibitem[{Devlin et~al.(2019)Devlin, Chang, Lee, and
  Toutanova}]{devlin-etal-2019-bert}
Devlin, J.; Chang, M.-W.; Lee, K.; and Toutanova, K. 2019.
\newblock {BERT}: Pre-training of Deep Bidirectional Transformers for Language
  Understanding.
\newblock In \emph{Proceedings of the 2019 Conference of the North {A}merican
  Chapter of the Association for Computational Linguistics: Human Language
  Technologies, Volume 1 (Long and Short Papers)}, 4171--4186. Minneapolis,
  Minnesota: Association for Computational Linguistics.

\bibitem[{Fader, Zettlemoyer, and Etzioni(2013)}]{fader-etal-2013-paraphrase}
Fader, A.; Zettlemoyer, L.; and Etzioni, O. 2013.
\newblock Paraphrase-Driven Learning for Open Question Answering.
\newblock In \emph{Proceedings of the 51st Annual Meeting of the Association
  for Computational Linguistics}, 1608--1618. Sofia, Bulgaria: Association for
  Computational Linguistics.

\bibitem[{Lan et~al.(2019)Lan, Chen, Goodman, Gimpel, Sharma, and
  Soricut}]{lan2019albert}
Lan, Z.; Chen, M.; Goodman, S.; Gimpel, K.; Sharma, P.; and Soricut, R. 2019.
\newblock Albert: A lite bert for self-supervised learning of language
  representations.
\newblock \emph{arXiv preprint arXiv:1909.11942} .

\bibitem[{Levesque, Davis, and Morgenstern(2012)}]{levesque2012winograd}
Levesque, H.; Davis, E.; and Morgenstern, L. 2012.
\newblock The winograd schema challenge.
\newblock In \emph{Thirteenth International Conference on the Principles of
  Knowledge Representation and Reasoning}. Citeseer.

\bibitem[{Lin et~al.(2019)Lin, Chen, Chen, and Ren}]{lin-etal-2019-kagnet}
Lin, B.~Y.; Chen, X.; Chen, J.; and Ren, X. 2019.
\newblock {K}ag{N}et: Knowledge-Aware Graph Networks for Commonsense Reasoning.
\newblock In \emph{Proceedings of the 2019 Conference on Empirical Methods in
  Natural Language Processing and the 9th International Joint Conference on
  Natural Language Processing (EMNLP-IJCNLP)}, 2829--2839. Hong Kong, China:
  Association for Computational Linguistics.

\bibitem[{Lin, Sun, and Han(2017)}]{lin-etal-2017-reasoning}
Lin, H.; Sun, L.; and Han, X. 2017.
\newblock Reasoning with Heterogeneous Knowledge for Commonsense Machine
  Comprehension.
\newblock In \emph{Proceedings of the 2017 Conference on Empirical Methods in
  Natural Language Processing}, 2032--2043. Copenhagen, Denmark: Association
  for Computational Linguistics.

\bibitem[{Liu et~al.(2020)Liu, Zhou, Zhao, Wang, Ju, Deng, and Wang}]{liu2020k}
Liu, W.; Zhou, P.; Zhao, Z.; Wang, Z.; Ju, Q.; Deng, H.; and Wang, P. 2020.
\newblock K-BERT: Enabling Language Representation with Knowledge Graph.
\newblock In \emph{Proceedings of the Thirty-Forth AAAI Conference on
  Artificial Intelligence}, 2901--2908.

\bibitem[{Liu et~al.(2019)Liu, Ott, Goyal, Du, Joshi, Chen, Levy, Lewis,
  Zettlemoyer, and Stoyanov}]{liu2019roberta}
Liu, Y.; Ott, M.; Goyal, N.; Du, J.; Joshi, M.; Chen, D.; Levy, O.; Lewis, M.;
  Zettlemoyer, L.; and Stoyanov, V. 2019.
\newblock Roberta: A robustly optimized bert pretraining approach.
\newblock \emph{arXiv preprint arXiv:1907.11692} .

\bibitem[{LoBue and Yates(2011)}]{lobue-yates-2011-types}
LoBue, P.; and Yates, A. 2011.
\newblock Types of Common-Sense Knowledge Needed for Recognizing Textual
  Entailment.
\newblock In \emph{Proceedings of the 49th Annual Meeting of the Association
  for Computational Linguistics: Human Language Technologies}, 329--334.
  Portland, Oregon, USA: Association for Computational Linguistics.

\bibitem[{Logan et~al.(2019)Logan, Liu, Peters, Gardner, and
  Singh}]{logan-etal-2019-baracks}
Logan, R.; Liu, N.~F.; Peters, M.~E.; Gardner, M.; and Singh, S. 2019.
\newblock {B}arack{'}s Wife Hillary: Using Knowledge Graphs for Fact-Aware
  Language Modeling.
\newblock In \emph{Proceedings of the 57th Annual Meeting of the Association
  for Computational Linguistics}, 5962--5971. Florence, Italy: Association for
  Computational Linguistics.

\bibitem[{Lv et~al.(2020)Lv, Guo, Xu, Tang, Duan, Gong, Shou, Jiang, Cao, and
  Hu}]{lv2020graph}
Lv, S.; Guo, D.; Xu, J.; Tang, D.; Duan, N.; Gong, M.; Shou, L.; Jiang, D.;
  Cao, G.; and Hu, S. 2020.
\newblock Graph-Based Reasoning over Heterogeneous External Knowledge for
  Commonsense Question Answering.
\newblock In \emph{Proceedings of the Thirty-Forth AAAI Conference on
  Artificial Intelligence}, 8449--8456.

\bibitem[{Ma et~al.(2019)Ma, Francis, Lu, Nyberg, and
  Oltramari}]{ma2019towards}
Ma, K.; Francis, J.; Lu, Q.; Nyberg, E.; and Oltramari, A. 2019.
\newblock Towards Generalizable Neuro-Symbolic Systems for Commonsense Question
  Answering.
\newblock In \emph{Proceedings of the First Workshop on Commonsense Inference
  in Natural Language Processing}, 22--32.

\bibitem[{Mihaylov and Frank(2018)}]{mihaylov-frank-2018-knowledgeable}
Mihaylov, T.; and Frank, A. 2018.
\newblock Knowledgeable Reader: Enhancing Cloze-Style Reading Comprehension
  with External Commonsense Knowledge.
\newblock In \emph{Proceedings of the 56th Annual Meeting of the Association
  for Computational Linguistics}, 821--832. Melbourne, Australia: Association
  for Computational Linguistics.

\bibitem[{Minsky(2000)}]{minsky2000commonsense}
Minsky, M. 2000.
\newblock Commonsense-based interfaces.
\newblock \emph{Communications of the ACM} 43(8): 66--73.

\bibitem[{Mintz et~al.(2009)Mintz, Bills, Snow, and
  Jurafsky}]{mintz-etal-2009-distant}
Mintz, M.; Bills, S.; Snow, R.; and Jurafsky, D. 2009.
\newblock Distant supervision for relation extraction without labeled data.
\newblock In \emph{Proceedings of the Joint Conference of the 47th Annual
  Meeting of the {ACL} and the 4th International Joint Conference on Natural
  Language Processing of the {AFNLP}}, 1003--1011. Suntec, Singapore:
  Association for Computational Linguistics.

\bibitem[{Pavlick et~al.(2015)Pavlick, Rastogi, Ganitkevitch, Van~Durme, and
  Callison-Burch}]{pavlick-etal-2015-ppdb}
Pavlick, E.; Rastogi, P.; Ganitkevitch, J.; Van~Durme, B.; and Callison-Burch,
  C. 2015.
\newblock {PPDB} 2.0: Better paraphrase ranking, fine-grained entailment
  relations, word embeddings, and style classification.
\newblock In \emph{Proceedings of the 53rd Annual Meeting of the Association
  for Computational Linguistics and the 7th International Joint Conference on
  Natural Language Processing}, 425--430. Beijing, China: Association for
  Computational Linguistics.

\bibitem[{Rahman and Ng(2012)}]{rahman-ng-2012-resolving}
Rahman, A.; and Ng, V. 2012.
\newblock Resolving Complex Cases of Definite Pronouns: The {W}inograd Schema
  Challenge.
\newblock In \emph{Proceedings of the 2012 Joint Conference on Empirical
  Methods in Natural Language Processing and Computational Natural Language
  Learning}, 777--789. Jeju Island, Korea: Association for Computational
  Linguistics.

\bibitem[{Rajani et~al.(2019)Rajani, McCann, Xiong, and
  Socher}]{rajani-etal-2019-explain}
Rajani, N.~F.; McCann, B.; Xiong, C.; and Socher, R. 2019.
\newblock Explain Yourself! Leveraging Language Models for Commonsense
  Reasoning.
\newblock In \emph{Proceedings of the 57th Annual Meeting of the Association
  for Computational Linguistics}, 4932--4942. Florence, Italy: Association for
  Computational Linguistics.

\bibitem[{Rajpurkar et~al.(2016)Rajpurkar, Zhang, Lopyrev, and
  Liang}]{rajpurkar-etal-2016-squad}
Rajpurkar, P.; Zhang, J.; Lopyrev, K.; and Liang, P. 2016.
\newblock {SQ}u{AD}: 100,000+ Questions for Machine Comprehension of Text.
\newblock In \emph{Proceedings of the 2016 Conference on Empirical Methods in
  Natural Language Processing}, 2383--2392. Austin, Texas: Association for
  Computational Linguistics.

\bibitem[{Robertson and Walker(1994)}]{robertson1994some}
Robertson, S.~E.; and Walker, S. 1994.
\newblock Some simple effective approximations to the 2-poisson model for
  probabilistic weighted retrieval.
\newblock In \emph{SIGIR’94}, 232--241. Springer.

\bibitem[{Sap et~al.(2019{\natexlab{a}})Sap, Le~Bras, Allaway, Bhagavatula,
  Lourie, Rashkin, Roof, Smith, and Choi}]{sap2019atomic}
Sap, M.; Le~Bras, R.; Allaway, E.; Bhagavatula, C.; Lourie, N.; Rashkin, H.;
  Roof, B.; Smith, N.~A.; and Choi, Y. 2019{\natexlab{a}}.
\newblock Atomic: An atlas of machine commonsense for if-then reasoning.
\newblock In \emph{Proceedings of the Thirty-Third AAAI Conference on
  Artificial Intelligence}, volume~33, 3027--3035.

\bibitem[{Sap et~al.(2019{\natexlab{b}})Sap, Rashkin, Chen, Le~Bras, and
  Choi}]{sap-etal-2019-social}
Sap, M.; Rashkin, H.; Chen, D.; Le~Bras, R.; and Choi, Y. 2019{\natexlab{b}}.
\newblock Social {IQ}a: Commonsense Reasoning about Social Interactions.
\newblock In \emph{Proceedings of the 2019 Conference on Empirical Methods in
  Natural Language Processing and the 9th International Joint Conference on
  Natural Language Processing (EMNLP-IJCNLP)}, 4463--4473. Hong Kong, China:
  Association for Computational Linguistics.

\bibitem[{Seo et~al.(2016)Seo, Kembhavi, Farhadi, and
  Hajishirzi}]{seo2016bidirectional}
Seo, M.; Kembhavi, A.; Farhadi, A.; and Hajishirzi, H. 2016.
\newblock Bidirectional attention flow for machine comprehension.
\newblock \emph{arXiv preprint arXiv:1611.01603} .

\bibitem[{Speer, Chin, and Havasi(2017)}]{speer2017conceptnet}
Speer, R.; Chin, J.; and Havasi, C. 2017.
\newblock ConceptNet 5.5: an open multilingual graph of general knowledge.
\newblock In \emph{Proceedings of the Thirty-First AAAI Conference on
  Artificial Intelligence}, 4444--4451.

\bibitem[{Talmor et~al.(2019)Talmor, Herzig, Lourie, and
  Berant}]{talmor2019commonsenseqa}
Talmor, A.; Herzig, J.; Lourie, N.; and Berant, J. 2019.
\newblock CommonsenseQA: A Question Answering Challenge Targeting Commonsense
  Knowledge.
\newblock In \emph{Proceedings of the 2019 Conference of the North American
  Chapter of the Association for Computational Linguistics: Human Language
  Technologies, Volume 1 (Long and Short Papers)}, 4149--4158.

\bibitem[{Wang and Jiang(2019)}]{wang-jiang-2019-explicit}
Wang, C.; and Jiang, H. 2019.
\newblock Explicit Utilization of General Knowledge in Machine Reading
  Comprehension.
\newblock In \emph{Proceedings of the 57th Annual Meeting of the Association
  for Computational Linguistics}, 2263--2272. Florence, Italy: Association for
  Computational Linguistics.

\bibitem[{Weissenborn, Ko{\v{c}}isk{\`y}, and
  Dyer(2017)}]{weissenborn2017dynamic}
Weissenborn, D.; Ko{\v{c}}isk{\`y}, T.; and Dyer, C. 2017.
\newblock Dynamic integration of background knowledge in neural NLU systems.
\newblock \emph{arXiv preprint arXiv:1706.02596} .

\bibitem[{Yang et~al.(2019{\natexlab{a}})Yang, Wang, Liu, Liu, Lyu, Wu, She,
  and Li}]{yang-etal-2019-enhancing-pre}
Yang, A.; Wang, Q.; Liu, J.; Liu, K.; Lyu, Y.; Wu, H.; She, Q.; and Li, S.
  2019{\natexlab{a}}.
\newblock Enhancing Pre-Trained Language Representations with Rich Knowledge
  for Machine Reading Comprehension.
\newblock In \emph{Proceedings of the 57th Annual Meeting of the Association
  for Computational Linguistics}, 2346--2357. Florence, Italy: Association for
  Computational Linguistics.

\bibitem[{Yang and Mitchell(2017)}]{yang-mitchell-2017-leveraging}
Yang, B.; and Mitchell, T. 2017.
\newblock Leveraging Knowledge Bases in {LSTM}s for Improving Machine Reading.
\newblock In \emph{Proceedings of the 55th Annual Meeting of the Association
  for Computational Linguistics}, 1436--1446. Vancouver, Canada: Association
  for Computational Linguistics.

\bibitem[{Yang et~al.(2019{\natexlab{b}})Yang, Dai, Yang, Carbonell,
  Salakhutdinov, and Le}]{yang2019xlnet}
Yang, Z.; Dai, Z.; Yang, Y.; Carbonell, J.; Salakhutdinov, R.~R.; and Le, Q.~V.
  2019{\natexlab{b}}.
\newblock Xlnet: Generalized autoregressive pretraining for language
  understanding.
\newblock In \emph{Advances in neural information processing systems},
  5753--5763.

\bibitem[{Zellers et~al.(2019)Zellers, Holtzman, Bisk, Farhadi, and
  Choi}]{zellers-etal-2019-hellaswag}
Zellers, R.; Holtzman, A.; Bisk, Y.; Farhadi, A.; and Choi, Y. 2019.
\newblock {H}ella{S}wag: Can a Machine Really Finish Your Sentence?
\newblock In \emph{Proceedings of the 57th Annual Meeting of the Association
  for Computational Linguistics}, 4791--4800. Florence, Italy: Association for
  Computational Linguistics.

\bibitem[{Zhang et~al.(2020)Zhang, Liu, Pan, Song, and Leung}]{zhang2020aser}
Zhang, H.; Liu, X.; Pan, H.; Song, Y.; and Leung, C. W.-K. 2020.
\newblock ASER: A large-scale eventuality knowledge graph.
\newblock In \emph{Proceedings of The Web Conference 2020}, 201--211.

\bibitem[{Zhang et~al.(2019{\natexlab{a}})Zhang, Song, Song, and
  Yu}]{zhang-etal-2019-knowledge}
Zhang, H.; Song, Y.; Song, Y.; and Yu, D. 2019{\natexlab{a}}.
\newblock Knowledge-aware Pronoun Coreference Resolution.
\newblock In \emph{Proceedings of the 57th Annual Meeting of the Association
  for Computational Linguistics}, 867--876. Florence, Italy: Association for
  Computational Linguistics.

\bibitem[{Zhang et~al.(2019{\natexlab{b}})Zhang, Han, Liu, Jiang, Sun, and
  Liu}]{zhang-etal-2019-ernie}
Zhang, Z.; Han, X.; Liu, Z.; Jiang, X.; Sun, M.; and Liu, Q.
  2019{\natexlab{b}}.
\newblock {ERNIE}: Enhanced Language Representation with Informative Entities.
\newblock In \emph{Proceedings of the 57th Annual Meeting of the Association
  for Computational Linguistics}, 1441--1451. Florence, Italy: Association for
  Computational Linguistics.

\bibitem[{Zhou et~al.(2018)Zhou, Young, Huang, Zhao, Xu, and
  Zhu}]{zhou2018commonsense}
Zhou, H.; Young, T.; Huang, M.; Zhao, H.; Xu, J.; and Zhu, X. 2018.
\newblock Commonsense Knowledge Aware Conversation Generation with Graph
  Attention.
\newblock In \emph{IJCAI}, 4623--4629.

\bibitem[{Zhou et~al.(2020)Zhou, Zhang, Cui, and Huang}]{zhou2020evaluating}
Zhou, X.; Zhang, Y.; Cui, L.; and Huang, D. 2020.
\newblock Evaluating Commonsense in Pre-Trained Language Models.
\newblock In \emph{Proceedings of the Thirty-Forth AAAI Conference on
  Artificial Intelligence}, 9733--9740.

\end{thebibliography}
\bibliographystyle{aaai21}

\end{document}